\documentclass[3p,times,twocolumn]{elsarticle}

\usepackage{color}
\usepackage{caption}
\usepackage{subcaption}
\usepackage{epstopdf}

\usepackage{longtable}

\usepackage{amssymb}
\usepackage{amsthm}

\usepackage[linesnumbered,ruled]{algorithm2e}
\usepackage{amsmath}
\usepackage{amssymb}
\usepackage{amsthm}
\usepackage{bm}
\usepackage{boxedminipage}
\usepackage{dsfont}
\usepackage{graphicx}
\usepackage{lipsum}
\usepackage{multirow}
\usepackage{paracol}
\usepackage{savesym}
\savesymbol{AND}
\usepackage{framed}

\theoremstyle{definition}
\newtheorem{theorem}{Theorem}
\theoremstyle{definition}
\newtheorem{lemma}[theorem]{Lemma}
\theoremstyle{definition}

\theoremstyle{definition}

\theoremstyle{definition}

\theoremstyle{definition}
\newtheorem{definition}[theorem]{Definition}
\theoremstyle{definition}
\newtheorem{proposition}[theorem]{Proposition}
\theoremstyle{definition}
\newtheorem{property}[theorem]{Property}
\theoremstyle{definition}
\newtheorem{example}[theorem]{Example}
\theoremstyle{definition}
\newtheorem{conjecture}[theorem]{Conjecture}
\theoremstyle{definition}
\newtheorem{remark}[theorem]{Remark}
\theoremstyle{definition}
\newtheorem{assumption}[theorem]{Assumption}
\theoremstyle{definition}
\newtheorem{corollary}[theorem]{Corollary}
\theoremstyle{definition}
\newtheorem{exercise}[theorem]{Exercise}
\theoremstyle{definition}
\newtheorem{simulation}[theorem]{Simulation}
\theoremstyle{definition}
\newtheorem{setting}{Setting}

 {\endMakeFramed}

\newenvironment{greenleftbar}{%
  \MakeFramed {\advance\hsize-\width \FrameRestore}}%
 {\endMakeFramed}

{\endMakeFramed}

\newenvironment{exercise-waku}
  {\begin{greenleftbar}\begin{exercise}}
  {\end{exercise}\end{greenleftbar}}
\newenvironment{simulation-waku}
  {\begin{greenleftbar}\begin{simulation}}
  {\end{simulation}\end{greenleftbar}}
\newenvironment{proposition-waku}
  {\begin{oframed}\begin{proposition}}
  {\end{proposition}\end{oframed}}
\newenvironment{definition-waku}
  {\begin{oframed}\begin{definition}}
  {\end{definition}\end{oframed}}
  \newenvironment{lemma-waku}
  {\begin{oframed}\begin{lemma}}
  {\end{lemma}\end{oframed}}
\newenvironment{remark-waku}
  {\begin{oframed}\begin{remark}}
  {\end{remark}\end{oframed}}
\newenvironment{theorem-waku}
  {\begin{oframed}\begin{theorem}}
  {\end{theorem}\end{oframed}}
\newenvironment{property-waku}
  {\begin{oframed}\begin{property}}
  {\end{property}\end{oframed}}
\newenvironment{corollary-waku}
  {\begin{oframed}\begin{corollary}}
  {\end{corollary}\end{oframed}}
  \newenvironment{example-waku}
  {\begin{oframed}\begin{example}}
  {\end{example}\end{oframed}}
  \newenvironment{conjecture-waku}
  {\begin{oframed}\begin{conjecture}}
  {\end{conjecture}\end{oframed}}
  \newenvironment{assumption-waku}
  {\begin{oframed}\begin{assumption}}
  {\end{assumption}\end{oframed}}
\newenvironment{setting-waku}
  {\begin{oframed}\begin{setting}}
  {\end{setting}\end{oframed}}

\newcommand{\vell}{{\bm{\ell}}}

\newcommand{\vs}{{\bm{s}}}

\newcommand{\vw}{{\bm{w}}}

\newcommand{\x}{{\bm{x}}}

\newcommand{\bR}{{\mathbb{R}}}

\newcommand{\vW}{{\bm{W}}}

\newcommand{\vthet}{{\bm{\theta}}}

\newcommand{\vPsi}{{\bm{\Psi}}}

 {\end{tabular}\end{flushleft}}

\newenvironment{tsaligned*}{\begin{equation*}\begin{aligned}}{\end{aligned}\end{equation*}}

\begin{document}

\begin{frontmatter}

  
  
  \title{A Study on Deep CNN Structures for Defect Detection From Laser Ultrasonic
  Visualization Testing Images}
  
  
  \author{Miya Nakajima\fnref{a1}\corref{cor1}}
  \author[a1]{Takahiro Saitoh}
  \author[a2]{Tsuyoshi Kato}
  
  \affiliation[a1]{organization={Graduate School of Science and Technology, Gunma University},
              addressline={Tenjin, Kiryu 376 8515, Japan}, 
              }
  \affiliation[a2]{organization={Faculty of Informatics, Gunma University},
              addressline={ Aramaki, Maebashi 371 8510, Japan}, 
              }
  \begin{abstract}
  The importance of ultrasonic nondestructive testing has been increasing in recent years, and there are high expectations for the potential of laser ultrasonic visualization testing, which combines laser ultrasonic testing with scattered wave visualization technology. Even if scattered waves are visualized, inspectors still need to carefully inspect the images. To automate this, this paper proposes a deep neural network for automatic defect detection and localization in LUVT images. To explore the structure of a neural network suitable to this task, we compared the LUVT image analysis problem with the generic object detection problem. Numerical experiments using real-world data from a SUS304 flat plate showed that the proposed method is more effective than the general object detection model in terms of prediction performance. We also show that the computational time required for prediction is faster than that of the general object detection model.
  \end{abstract}
  
  
  
\end{frontmatter}

\section{Introduction}
Nondestructive testing has become increasingly important in recent years. 
Ultrasonic nondestructive testing is the most widely used nondestructive testing method in engineering applications.
Ultrasonic nondestructive testing uses an ultrasonic probe to inject ultrasonic waves into an inspection target and detect internal defects from the scattered waves obtained.
In general, it is difficult to determine details such as the location of defects only by viewing the measured scattered waveforms.
Therefore, the location of defects is identified by inverse analysis methods such as aperture synthesis method~\cite{Doctor86ndtint}, inverse scattering analysis~\cite{kitahara02wm}, and time reversal method~\cite{Fink92tuf}. 
Meanwhile, the laser ultrasonic method, which enables non-contact inspection of structures and materials, has recently been developed and applied in the field of nondestructive testing.
In particular, the laser ultrasonic visualization test (LUVT)~\cite{Kohler02usonics,Yashiro08ndtei}, which is an attempt to visualize ultrasonic propagation on the surface of a specimen, has the advantage that defects inside structures and materials can be detected relatively easily without the need for a skilled inspector. 
So far, applications to metallic materials such as steel and aluminum, materials that exhibit acoustic anisotropy properties such as CFRP~\cite{SaitohMori19insight}, and the nuclear and aerospace industries~\cite{Virkkunen2014ecndt,Siljama2021,Virkkunen2021virtual} have been reported.
Despite the impressive developments in ultrasound technology, it is difficult to ignore the shortage of engineers. 
Even with the evolution of structural and material inspection equipment and ultrasonic visualization technology, human inspectors must still make the judgment on the presence or absence of defects for a huge amount of specimens, and the shortage of engineers, which has already crept up on us in the near future, must be urgently addressed. 

On the other hand, there are growing attempts to automate ultrasonic nondestructive testing through machine learning. 
Prior to the advent of deep neural nets, the predictive models used to automate ultrasonic nondestructive testing were low-capacity models involving shallow neural nets~\cite{Masnata96ndtei,Sambath10jnde} or SVMs~\cite{Huifang20ndtei,xiaokai19ultras}. 
In the 2010s, models with huge capacity, such as deep neural networks, have successively revolutionized a variety of fields, and deep convolutional neural networks have risen to the position of de facto standard in the field of image recognition~\cite{MingxingTan-cvpr2020,ItoTomIizHirKat16a,HenKat-tom22a}.
In response to this fact, a number of attempts have been made to use deep models for ultrasonic non-destructive testing~\cite{MENG2017,Munir18jmst,Pyle20tuffc,Koskinen21jnde,Siljama2021,Boikov21sym}. 
Unlike other machine learning applications, ultrasonic non-destructive testing using huge-capacity machine learning models has a serious problem, which is that the positive examples for training predictive models tend to be sparse, and the training datasets cannot be made large. 
To achieve better prediction performance for unseen data, it is known that careful selection of a model with a capacity commensurate with the size of the available data set is important\cite{JianxingYe18sensors,HasTibFri-book09a,TajHir-sdm21a}. 
The task of defect detection from LUVT images addressed in this study is highly related to the task of object detection in generic image recognition.
However, for the reasons mentioned above, there is concern that the direct use of models that have been successfully used for generic object detection in the field of image recognition may be at risk of overfitting. 

In this study, we re-designed the deep model structure to fit the task of defect detection from LUVT images. 
First of all, we examined the differences between the LUVT image analysis task and the generic object detection task. 
We then identified which modules of a typical state-of-the-art object detection model are necessary for the LUVT image analysis task and which modules are unnecessary.
We finally obtained a slimmer framework by eliminating unnecessary modules for LUVT image analysis.
This paper reports the experimental results of applying the resultant slimmer framework to real LUVT image data and comparing it with the state-of-the-art object detection model.

\begin{figure}[t]
    \centering
      \includegraphics[width=5cm]{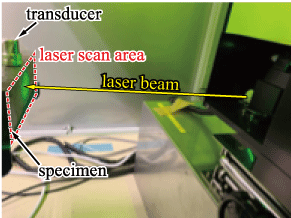}
      \caption{LUVT device. }
      \label{fig:laser}
    \end{figure}
\begin{figure}[t]
    \centering
    \begin{tabular}{ll}
    (a) & (b)
    \\
    \includegraphics[width=3.8cm]{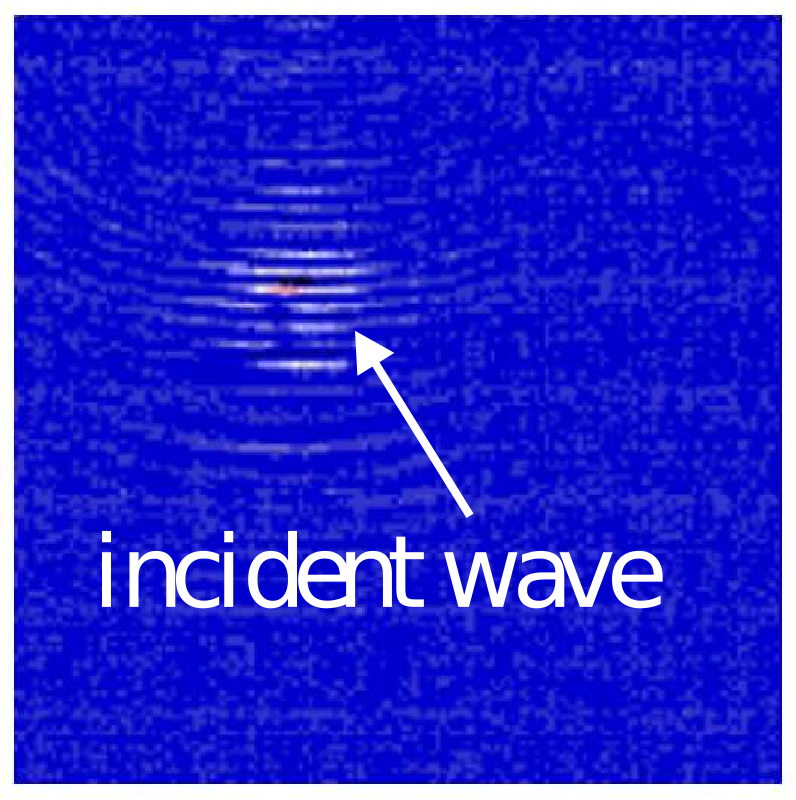}
       &
    \includegraphics[width=3.8cm]{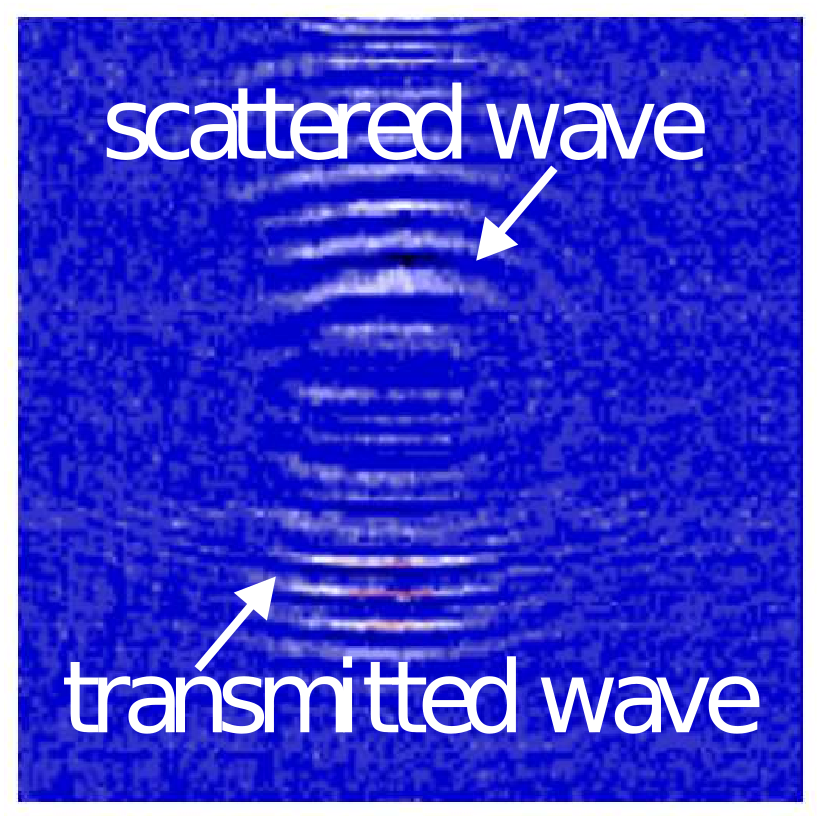}
    \\
    \end{tabular}
    \caption{Examples of LUVT images for SUS304 flat plates. (a) without defects. (b) with defects. \label{fig:example}}
  \end{figure}

\section{What is LUVT?}
In this section, a brief summary of LUVT is provided for the sake of explanation in the subsequent sections. 
In non-destructive inspection using LUVT, a laser is irradiated onto the target structure or material as shown in Figure~\ref{fig:laser}.
At this time, ultrasonic waves are generated at the laser irradiation point. 
This ultrasonic wave is then sent to a predetermined receiving point, where the ultrasonic waves are received at a predetermined receiving point which is an ultrasonic probe. 
The ultrasonic probe receives the ultrasonic waves by interchanging the transmitter and receiver points using the conflict theorem. 
The receiver point is then the ultrasonic wave source.
Interchanging the receiving point and the transmitting point produces the pseudo ultrasonic waveform transmitted from the ultrasonic probe at the receiving point. 
Therein, the transmitting point is the laser irradiation point.
In other words, when this operation is performed using the conflict theorem, the apparent sending point is the ultrasonic probe and the receiving point is the laser point, which is interchanged with the original setup.
Therefore, if the ultrasonic probe is fixed and the laser is densely scanned over a suitable region of the specimen, and then the conflict theorem is applied to all combinations of the ultrasonic probe and all laser irradiation points, a pseudo-time history image of ultrasonic propagation in the laser scan region emitted from the ultrasonic probe can be obtained.
Figure~\ref{fig:example} shows two examples of LUVT image obtained for a SUS304 flat plate, where Figure~\ref{fig:example}(a) is an LUVT image without defects, and Figure~\ref{fig:example}(b) is an LUVT image with a defect.
As shown in Figure~\ref{fig:example}(a), when there is no defect, the incident ultrasonic waves from the ultrasonic probe located above propagate directly without being scattered. When there is a defect, scattered waves from the defect are generated as shown in Figure~\ref{fig:example}(b). 
An advantage of LUVT over other ultrasonic tests is that even a non-expert engineer can easily see the scattered waves from defects and judge the presence or absence of defects in the specimen. 
We developed a deep neural network model that estimate the presence and location of defects based on the generation of such scattered waves.

\section{Characteristics of LUVT Image Analysis}
It has been more than a decade since deep learning was introduced to the field of computer science as a technique for detecting objects in images, and its performance has evolved at a dizzying pace~\cite{Dosovitskiy21-iclr,MingxingTan-cvpr2020,Henmi21icip}.
Rather than blindly trying the generic object detection techniques, a prior thorough understanding of the differences between generic object detection and the task in this study leads to development of a better machine learning algorithm.
This section provides a discussion on the comparison of LUVT image analysis and generic object detection.

\textbf{Dataset: }
When developing machine learning models, it is important to consider not only the power of the input to predict the output, but also the size of the data set available to train the model.
If the task is simpler and the dataset is larger, it is possible to train larger and more complex models and expect higher generalization performance. 
Meanwhile, if the dataset is small, large models will cause overfitting and reduce the generalization ability. 
In many generic object detection methods, data for training can be collected simply by taking pictures with a camera, making it relatively easy to acquire so-called \emph{big data} as long as the annotation cost is paid. 
In contrast to the generic object detection, collecting LUVT data requires a time-consuming and costly procedure, such as manufacturing an inspection target with internal defects, irradiating it with laser ultrasonic using a laser ultrasonic visualization system (Figure~\ref{fig:laser}), and performing the imaging process. 
The size of the dataset used for training, therefore, cannot be scaled easily due to the high cost paid for this data collection process.

\textbf{Number of detections: }
When benchmarking generic object detection techniques, all objects in each input image are required to be enumerated.
For example, in the self-driving task, it is necessary to detect every pedestrian because missing a pedestrian can lead directly to a serious accident.
In contrast, the LUVT task does not claim enumeration of all defects in a specimen, because the importance for the non-destructive testing lies in whether there are no defects or whether there are one or more defects.

\textbf{Number of categories: }
Deep learning evolved with multi-category classification as its most fundamental task.
Most generic object detection models incorporate this trend and are equipped with a module for multi-category classification.
In fact, many generic object detection tasks require categorization of detected objects.
For example, in the case of object detection from an in-vehicle camera, the object is classified into many categories including a pedestrian, a bicycle, a traffuc sign, and a bus. 
In the LUVT task, it is possible to classify a defect into one of types such as a crack and a cavity, although it is not necessary. 
Therefore, in this study, we decided not to add a module for categorization to the model framework.

\textbf{Background: } 
The backgrounds for generic object detection are often diverse.
For example, in the case of vehicle-mounted camera image analysis, the background may be an urban area, a rural landscape, daytime, or dusk.
The background of LUVT images is approximately uniform, although there is a lot of signal noise that is too strong to distinguish from the true scattered waves. 

\textbf{Bounding box: } 
Generic object detection is required to accurately estimate the rectangle surrounding the object, referred to as the bounding box.
For example, after detecting the bounding box of a pedestrian, pose estimation can be performed subsequently to predict the next action of the pedestrian. 
For LUVT, once the approximate center of the defect is known, it is not necessary to go as far as the bounding box, since the location can be examined in some other way if necessary.

\textbf{Preferable prediction performance: }
Depending on the application, generic object detection is usually required to reduce both false positives and false negatives at the same time.
For example, in the case of automatic driving, if a driver inadvertently thinks there is a pedestrian crossing the road and brakes suddenly, even though there is no pedestrian crossing, the car may be rear-ended. 
Meanwhile, overlooking a pedestrian crossing a road can cause a catastrophe.
For nondestructive testing of structures and materials, the reduction of false negatives is particularly important.
Even if a defect in a structure or material is detected incorrectly by LUVT, it can be corrected by the inspector's visual confirmation, so it does not cause serious losses.
In contrast, missing defects can cause significant losses in the future.

\begin{figure*}[t]
  \centering
  \begin{tabular}{ll}
  (a) Typical object detector
  &
  (b) Proposed model
  \\
  \\
  \includegraphics[height=4cm]{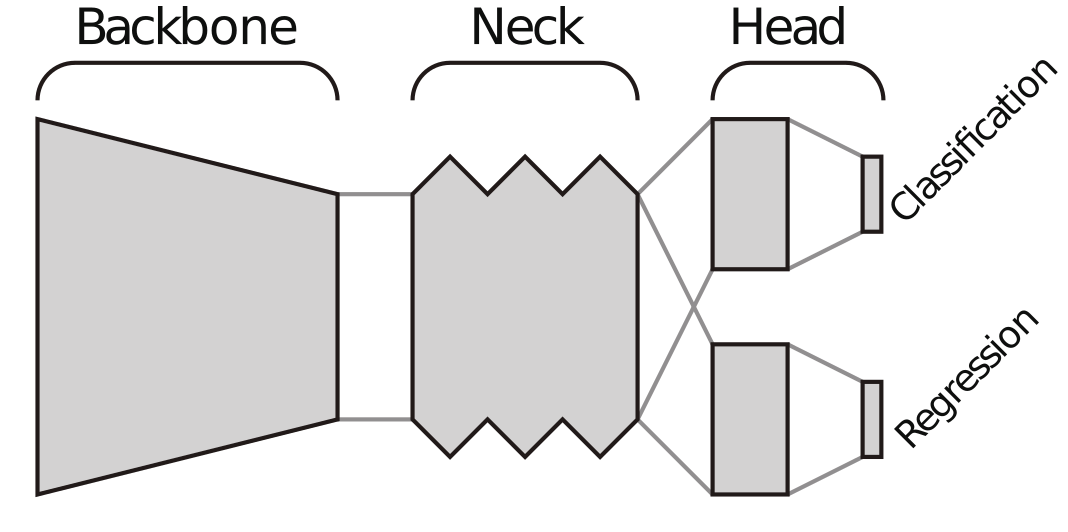}
  &
  \includegraphics[height=4cm]{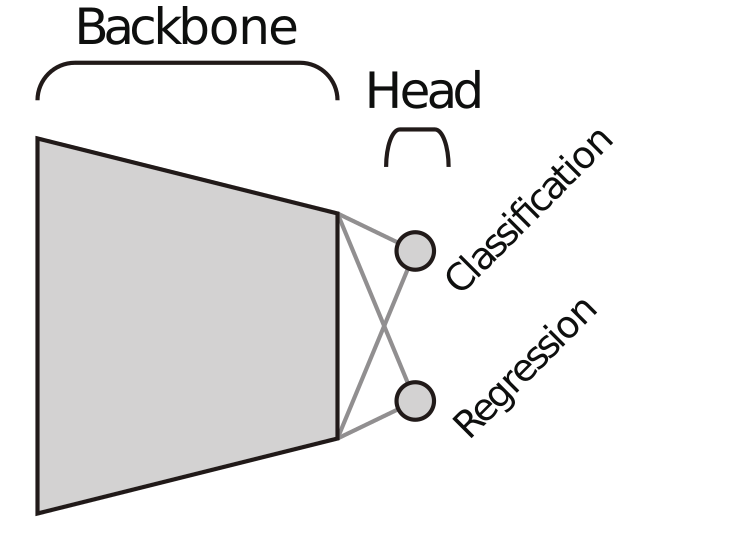}
  \end{tabular}
  \caption{Comparison of model structures. }
  \label{fig:mdl}
\end{figure*}

\section{Network Structures}
\label{s:model}
What structure is appropriate for a nerual network to detect defects in LUVT images is discussed in this section.

\textbf{Transformers: } 
Recently, the Vision Transformer (ViT)~\cite{Dosovitskiy21-iclr} has attracted attention as an alternative image classifier to convolution-based networks, and the Swin Transformer~\cite{Liu21-iccv,Liu2021} and DETR~\cite{Carion20-eccv,caron2021emerging} have emerged as ViT-based object detectors.
However, Transformer-based models require huge datasets for training and are considered unsuitable for the task of LUVT image analysis, where available datasets are limited~\cite{vaswani2017attention}.

\textbf{Convolutional networks: } 
Many convolutional network-based object detection algorithms have been developed before the emergence of ViT. 
There are two types of object detectors: two-stage and one-stage approaches.
The two-stage approach first estimates candidate regions and then removes non-target objects from the candidate regions~\cite{Girshick15-iccv,Malbog19-icetas}. 
Two-stage approaches are highly accurate but computationally expensive.
In contrast, the one-stage approach solves the issue of heavy computational complexity by combining the pipeline into a single network~\cite{liu2016ssd,Redmon-cvpr2016,XuemeiXie17-icvip}. 
Furthermore, the accuracy of the one-stage approach has already been comparable to the two-stage approach.

A majority of the high-performance one-step approaches employ the model structure shown in Figure~\ref{fig:mdl}(a)~\cite{MingxingTan-cvpr2020}. 
The model structure consists of three modules: the backbone, the neck, and the head.
Given an image input, the classification head outputs predictions of object categories, and the regression head outputs predictions of object locations.
The backbone extracts a feature map from the image.
The neck contains a multi-scaling module to avoid missing small objects.

\textbf{Proposed models: } 
Based on the characteristics of LUVT images described so far, and after the considerations described below, we re-considered simple models having emerged at the dawn of deep learning to develop an automatic defect detection method for LUVT~\cite{AlexNet2012}.
The requirements for LUVT are simpler than those for general object detection tasks, although the set of available data for training is small. 
To gain the usefulness of the prediction model, the model structures and the tasks are in need to be selected carefully. 
Consequently, the goal of this study was designed not to categorize, but only to predict the presence or absence of defects. 
The proposed model detects only one defect and predicts only the center location of that defect, not the bounding box. 
In such a task setting, the state-of-the-art object detection model has many unnecessary modules.
Therefore, in this study, we decided to eliminate the unnecessary modules, leaving the backbone that performs feature extraction.
Usually, non-destructive ultrasonic testing can detect defects of the same size as the wavelength of the ultrasonic waves being handled.
Although LUVT has a wide range of applications, this study is a basic investigation for detecting defects in metallic materials such as steel and aluminum, and aims to detect defects with a length of several millimeters. 
Defects that are sufficiently small compared to the wavelength of ultrasonic waves are ignored because ultrasonic waves penetrate the defect and no scattered waves are generated. 
Therefore, we did not aim to detect such small scratches in this study, which allows us to eliminate the neck module. 
The above considerations led to a simple model structure as shown in Figure~\ref{fig:mdl}(b).

\begin{figure}[t]
\centering
  \includegraphics[width=8cm]{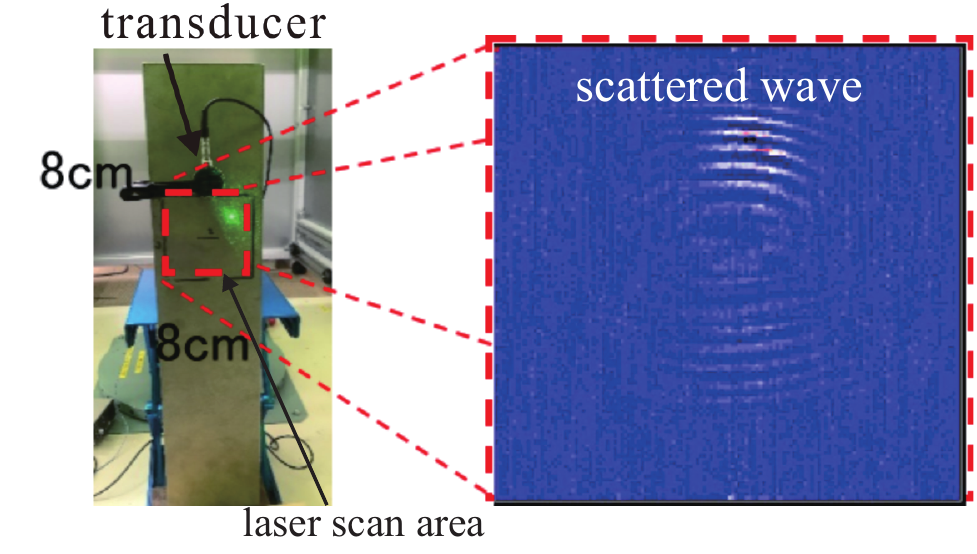}
  \caption{LUVT applied to SUS304 plain. }
  \label{fig:setup}
\end{figure}

\begin{table*}[t]
  \caption{Dataset.}
  \label{tab:exp-dataset}
  \centering
    \begin{tabular}{|c|c|c|c|c|c|c|} \hline 
      Series & Defect Position & \# defects & \# non-defects \\ 
      \hline \hline
      1  & No defect  & 0   & 582 \\
      2  & Center     & 219 & 281 \\
      3  & Right      & 272 & 228 \\
      4  & Left       & 271 & 229 \\
      5  & Upper        & 179 & 343 \\
      6  & Upper right  & 180 & 342 \\
      7  & Upper left   & 180 & 343 \\
      8  & Lower        & 157 & 364 \\
      9  & Lower right  & 158 & 364 \\
      10 & Lower left   & 159 & 363 \\
      \hline
    \end{tabular}
\end{table*}

\section{Dataset}
\label{s:dataset}

This section describes how a dataset for machine learning and performance assessment was prepared for the experiments reported in Section~\ref{s:exp}. 
The specimens were SUS304 flat plates with a thickness of 2 mm as shown in Figure~\ref{fig:setup}. 
The laser scan area was a square area of $8\text{cm}\times 8\text{cm}$, as shown by the red dotted line in Figure~\ref{fig:setup}.
The laser irradiation distance was approximately 59 cm. 
The laser irradiation pitch and the number of irradiated points in the area, respectively, were 0.501mm $\times$ 0.501mm and 159$\times$161=25,599 points for the horizontal and vertical directions. 
The ultrasonic probe was placed above the dotted line.
The LUVT principle follows the conflict theorem, which leads to the fact that the incident ultrasound in the LUVT image is incident from above the image.
The laser scan is performed over the entire area of this dotted box, allowing the defect detection within this area. 
The defects on the SUS304 flat plate were artificial defects, the shape of which is a slit of 2 cm in length, 1 mm in opening width, and 1 mm in depth. 
Hereinafter, the slits are simply referred to as defects. 
Nine positions were selected and a split was made in each position as shown in Table~\ref{tab:exp-dataset}. 
A surface wave probe with a center frequency of 2 MHz was used as the ultrasonic probe. 
Under these LUVT conditions, several SUS304 plates without and with defects were irradiated with lasers, and the ultrasonic propagation at each time was imaged by LUVT.

When LUVT is applied to a sample with defects, the incident ultrasonic waves are scattered by the defects at a certain time, generating scattered waves, as shown in Figure~\ref{fig:setup}.
These images consist of approximately 520 steps, which are the time history of the ultrasonic waves generated in the SUS304 plate from the ultrasonic probe at the top of the image until the incident wave finally reaches the bottom of the laser scan area.

In order to learn prediction models and to evaluate the prediction performance, it is necessary to assign the true binary class label to each image used for training and performance evaluation. 
In this study, we labeled the images between the time the ultrasonic wave hits the defect and the time when the wavefront of the scattered wave generated by the defect is no longer visible as the annotated label for the presence of a defect.
The other images were assigned the annotated label of no defect.
In addition to the presence/absence of defects, the location information of each defect was also annotated. 
Thus, an LUVT image dataset was organized as shown in Table~\ref{tab:exp-dataset}. 
To evaluate the performance of the prediction model, the dataset was divided into three exclusive subsets called training dataset, validation dataset, and test dataset.
The training data were used for training the prediction model.
The weight parameters were recorded for each epoch, and the weights with the best prediction performance on the validation dataset were selected as the learning results.
The selected weights were used to evaluate the performance on the testing data. 
The dataset consists of ten image series. 
Six of the ten series were used for training, two were for validation, and the remaining two were for testing.

\section{Learning Algorithm}
\label{s:train}
In this study, a predictor was developed to predict the presence or absence of a defect in a SUS304 flat plate, as well as to estimate the location of the defect.
The network shown in Figure~\ref{fig:mdl}(b) contains weight parameters to be adjusted for both the backbone and the head.
The values of the weight parameters are automatically determined from the training dataset.
The training of a neural network is usually done by minimizing an error function also known as an empirical risk function. 
The error function employed in this study consists of the sum of the error term for the classification task and the error term for the regression task.
The error term for the classification task was the average of the cross-entropy loss for each training data.
The cross-entropy loss is large when the output of the discriminative head is inconsistent with the actual presence or absence of defects.
The error term for the regression task was the mean squared error of the predicted positions for the LUVT images containing a defect. 

Learning is done by standard mini-batch gradient descent. 
Denote by $\vPsi(I;\vthet):=\left[\Psi_{1}(I;\vthet), \dots, \Psi_{d-1}(I;\vthet), 1 \right]^\top\in\bR^{d}$ the feature map extracted by the backbone module from an input image $I$, where $\vthet$ is the weight parameter included in the backbone. 
The classification head and the regression head use
$\vW^{\text{c}}:=\left[ \vw_{1,\text{c}}, \vw_{2,\text{c}} \right]$ and
$\vW^{\text{r}}:=\left[ \vw_{,\text{x,r}}, \vw_{\text{y,r}} \right]$, respectively, 
to output two-dimensional signals  
$(\vW^{\text{c}})^\top\vPsi(I\,;\,\vthet)$
and
$(\vW^{\text{r}})^\top\vPsi(I\,;\,\vthet)$.

Let $N$ be the number of training images.
Let $I^{(1)},\dots,I^{(N)}$ be the training images.
Each image $I^{(n)}$ is given a class label $c^{(n)}\in\{1,2\}$.
Let $N_{+}$ be the number of training images with defects and $N_{-}:=N-N_{+}$ be the number of training images without defects.
The training image order is arranged as $c^{(1)}=\dots=c^{(N_{+})}=1$, $c^{(N_{+}+1)}=\dots=c^{(N)}=2$. 
The weight parameters of the developed deep model are $(\vthet,\vW^{\text{c}},\vW^{\text{r}})$, and these values are adjusted to minimize the error function.
The error function
$R(\vthet,\vW^{\text{c}},\vW^{\text{r}})$ is a linear combination of the error term for classification and the error term for regression:
{\small
\begin{multline}\label{eq:multitaskloss}
  \hspace{-1mm} R(\vthet,\vW^{\text{c}},\vW^{\text{r}})
    :=
    \frac{1}{N}\sum_{n=1}^{N}\text{CE}((\vW^{\text{c}})^\top\vPsi(I^{(n)}\,;\,\vthet)\,;\,c^{(n)})
    \\
    \qquad
    +
    \frac{1}{N_{+}}\sum_{n=1}^{N_{+}}
    \lVert
    (\vW^{\text{r}})^\top\vPsi(I^{(n)}\,;\,\vthet)-\vell^{(n)}\rVert^{2}. 
\end{multline}
}
However, $\vell^{(n)}:=\left[x^{(n)},y^{(n)}\right]^{\top}$ is
$(x,y)$ coordinates of the defect location given by the annotation.
$\text{CE}(\cdot;c):\bR^{2}\to\bR$ denotes the cross-entropy loss function: 
\begin{align}
    \text{CE}(\vs\,;\,c) := \log\left(\sum_{j=1}^{2}\exp(s_{j}-s_{c})\right). 
\end{align}
By minimizing the error function $R$, the parameters of the classification and regression heads can be optimized, and at the same time the backbone for feature extraction can be tuned to be compatible with defect detection and localization. 

The size of the LUVT input image was set to $224\times 224$. 
The data were augmented by Resize, Crop, Shift, Scale, Rotate, and transforming Brightness, Contrast, and Hue-Saturation-Value. 
Normal noise was added for a countermeasure against noise.
A saturation transformation was also added to draw attention to the shape of the waveform.

Minimizing the error function $R$ is a convex problem if the backbone weights $\vthet$ are fixed to pre-trained values such as ImageNet and only $\vW^{\text{c}}$ and $\vW^{\text{r}}$ are adjusted, so optimization is easy~\cite{TajHenKat-ieice22a,KatHir-acml19a}.
Instead, if the backbone is to be optimized simultaneously, the optimization scheme must be carefully chosen.
Preliminary experiments showed that learning algorithm for the proposed model runs stably in the following settings.
Adam is used as the optimization method. 
The batch size is set to $256$. The initial learning rate is set to $10^{-3}$ and the model is trained for 30 epochs. 
In the first ten epochs, only the classification and regression heads are trained. 
From the 11th epoch to the 20th epoch, the learning rate is changed down to $10^{-4}$ and the tail blocks of the backbone are also updated. 
In the last ten epochs, the entire model is trained.

\begin{figure*}[t]
  \centering
  \begin{tabular}{l}
  \includegraphics[width=1.0\linewidth]{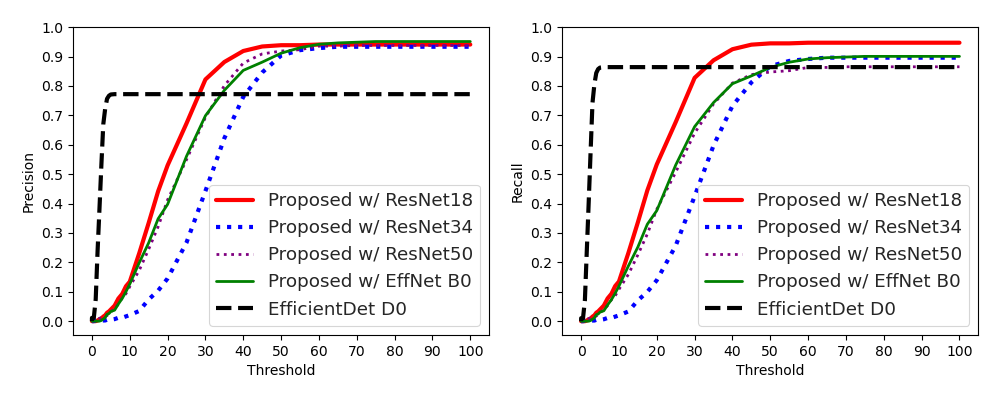}
  \end{tabular}
  \caption{Prediction performance. }
  \label{fig:acc}
\end{figure*}

\begin{figure*}[t!]
    \centering
    \begin{tabular}{ll}
      \\ \\
      (a) & (b)
      \\
      \includegraphics[width=5cm]{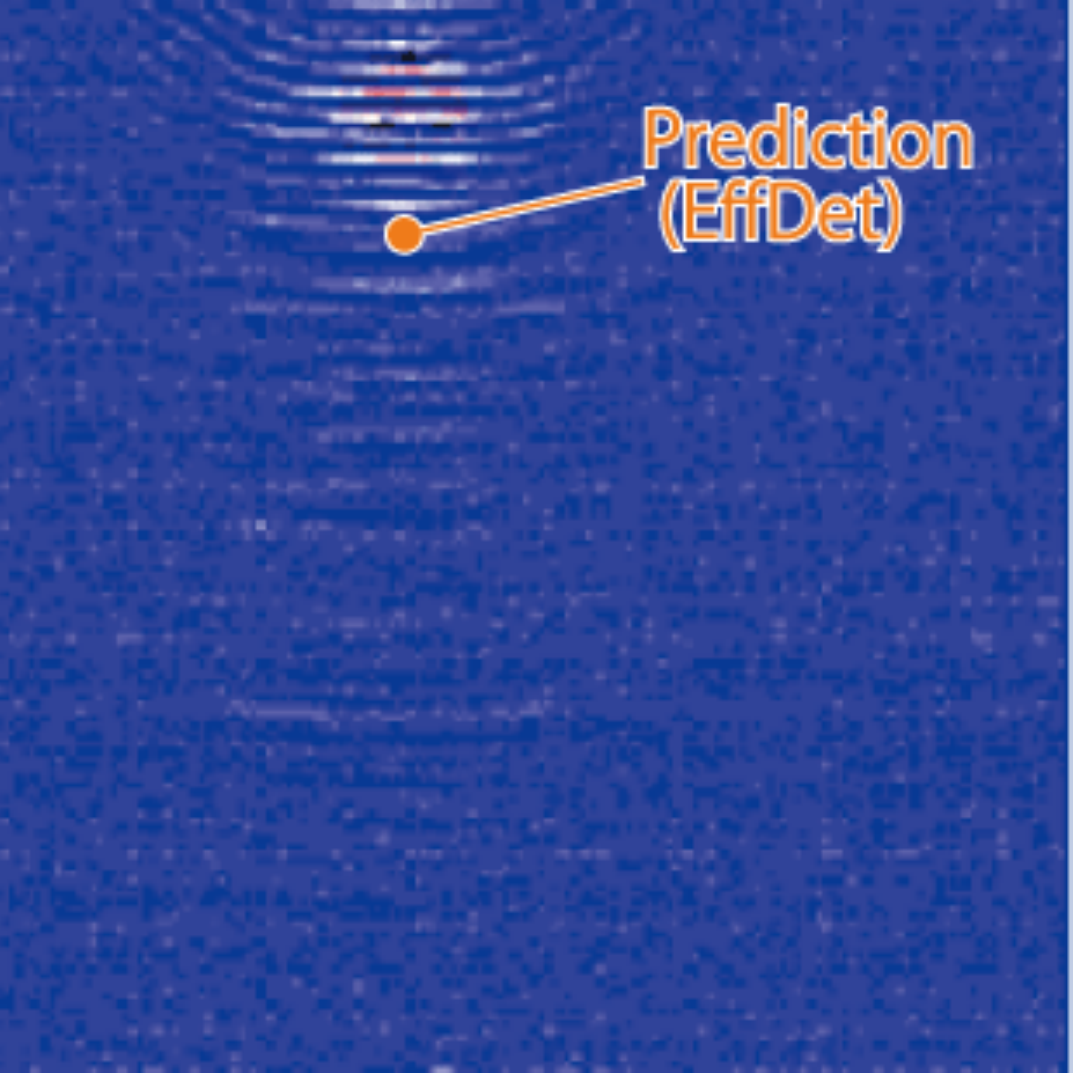}
      &
      \includegraphics[width=5cm]{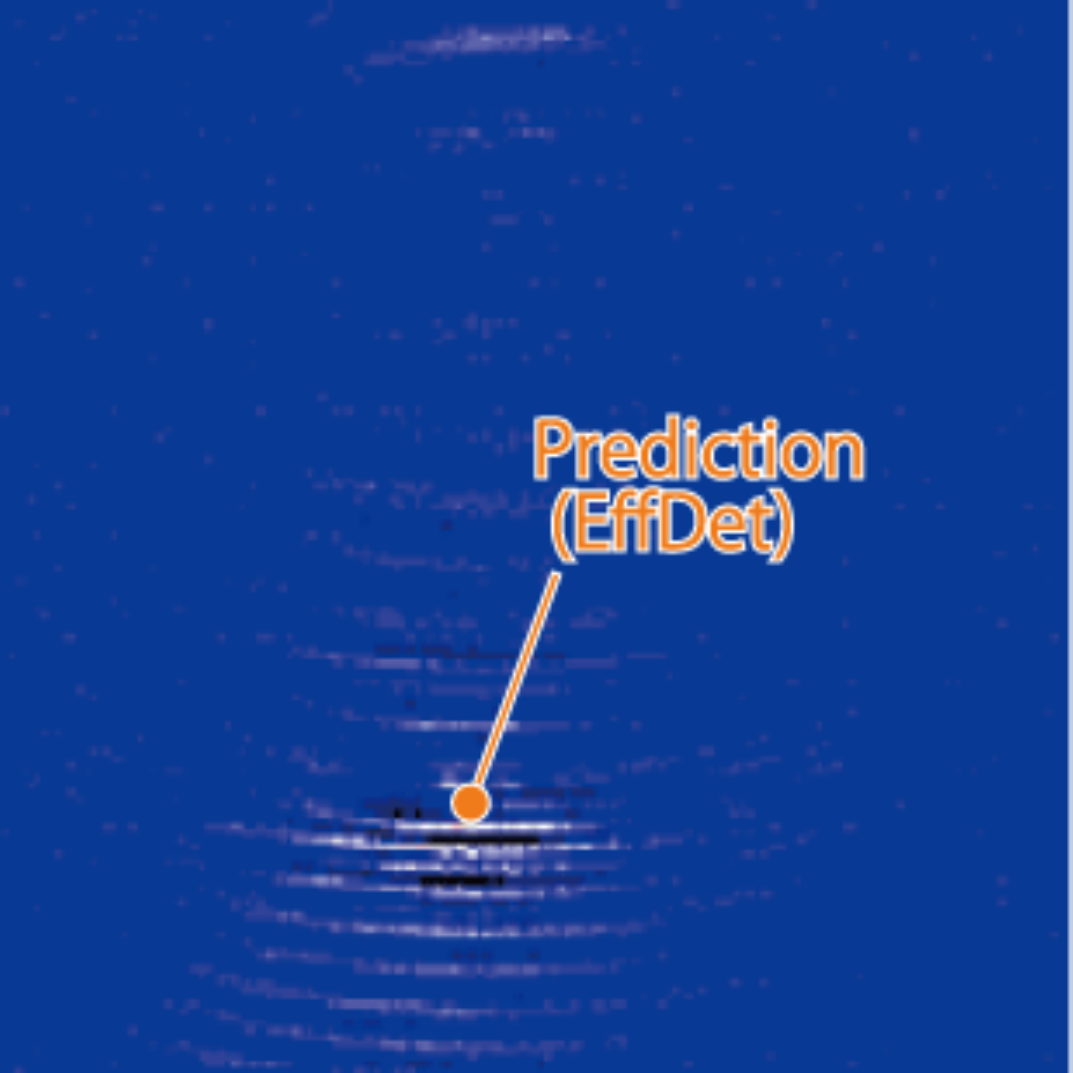}
      \\ \\
      (c) & (d)
      \\
      \includegraphics[width=5cm]{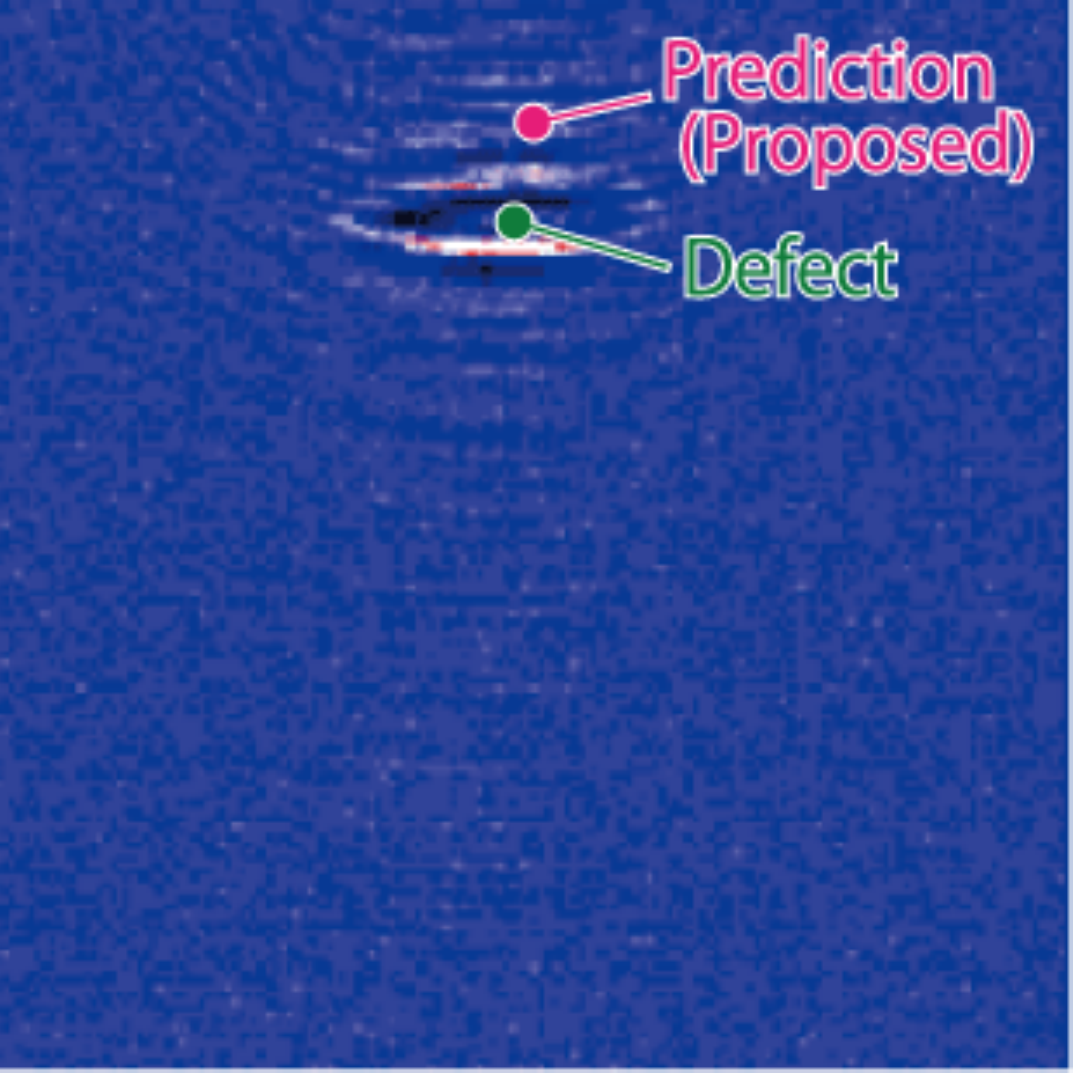}
      &
      \includegraphics[width=5cm]{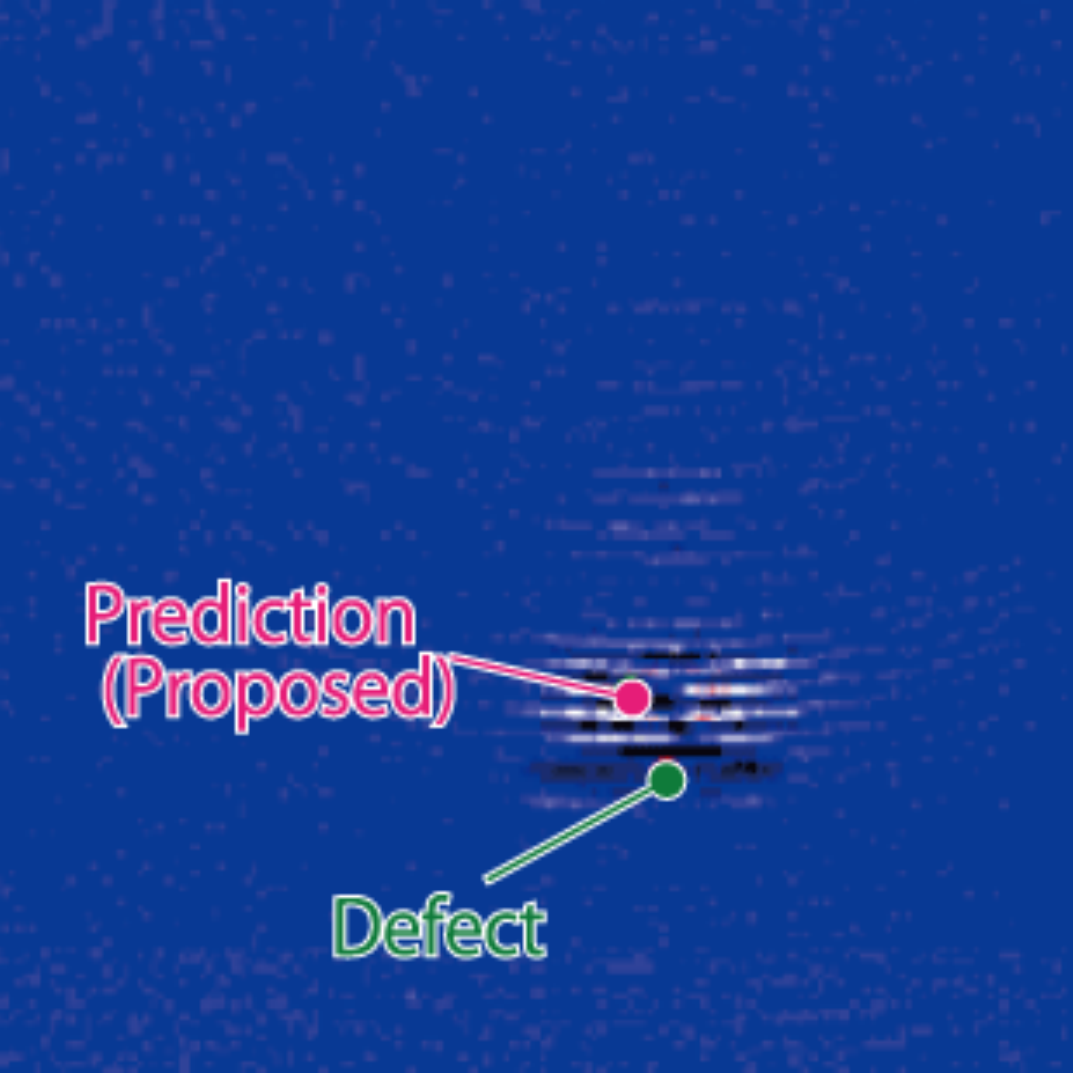}
      \\ \\
      (e) & (f)
      \\
      \includegraphics[width=5cm]{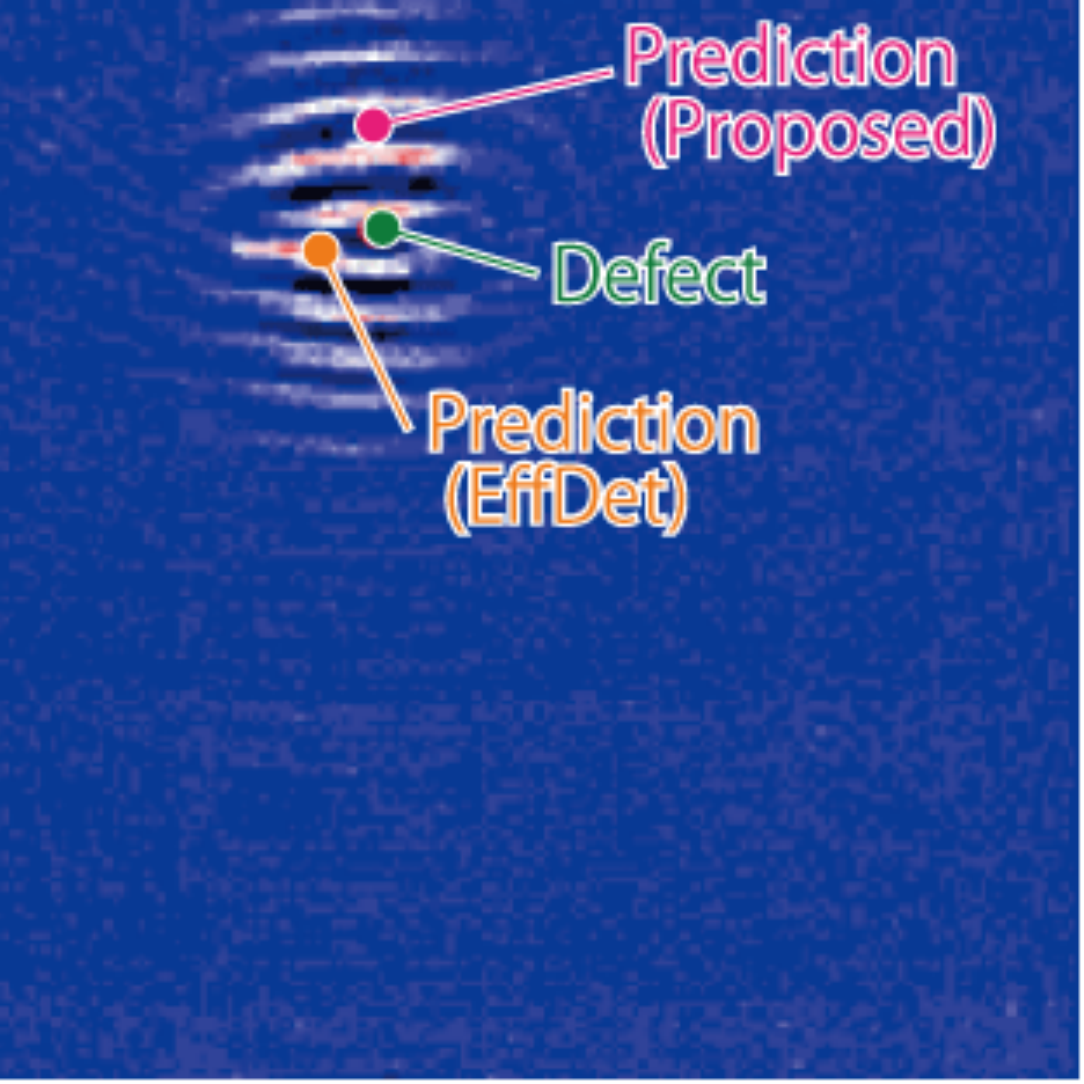}
      &
      \includegraphics[width=5cm]{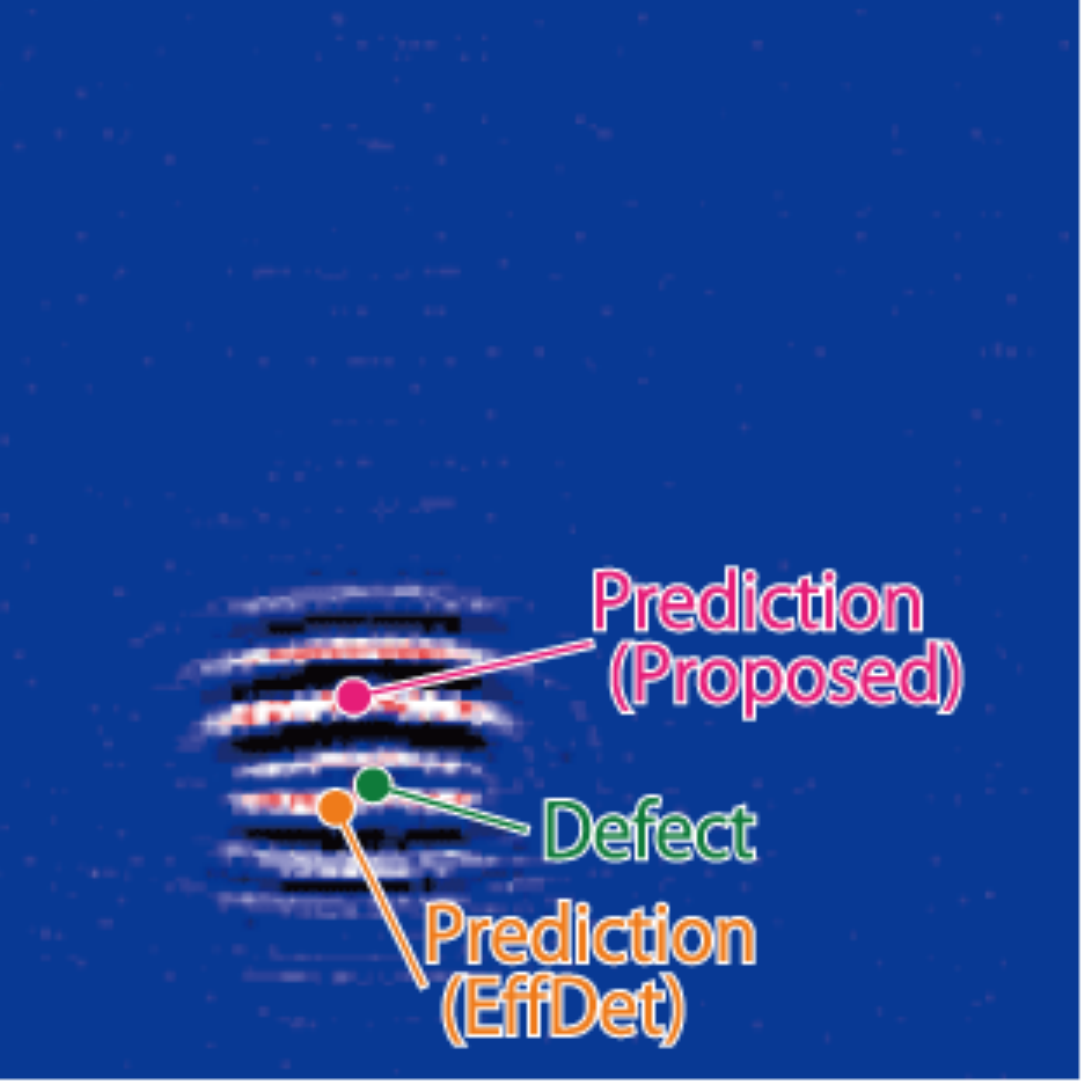}
    \end{tabular}
    \caption{Examples of prediction results. Proposed and EffDet are the abbreviations of the proposed method and EfficientDet, respectively. }
    \label{fig:predimg}
  \end{figure*}

\begin{table}[t]
  \caption{Capacity of models used for the experiments}
  \label{tab:exp-mem}
\centering
    \begin{tabular}{|c|c|c|c|} \hline 
      Backbone & Parameters  \\ \hline
      Proposed w/ ResNet18  & 11M   \\
      Proposed w/ ResNet34  & 21M   \\
      Proposed w/ ResNet50  & 23M   \\
      Proposed w/ EffNet B0  & 4M     \\
      EfficientDet D0  & 3M     \\
      \hline
    \end{tabular}
\end{table}

\begin{table}[t]
  \caption{System configurations of the computers used for the experiments. }
  \label{tab:exp-spec-a}
  \centering
    \begin{tabular}{|c|c|c|c|} \hline 
      \multicolumn{2}{|c|}{(a) GPU machine}  \\ \hline
      GPU      & RTX 3090       \\
      CPU      & Core i9 10850K \\
      Memory   & 64GB           \\
      \hline
    \end{tabular}

  \centering
    \begin{tabular}{|c|c|c|c|} \hline 
      \multicolumn{2}{|c|}{(b) CPU machine}  \\ \hline
      GPU      & None       \\
      CPU      & Core i9 10850K \\
      Memory  & 64GB           \\
      \hline
    \end{tabular}
\end{table}

\begin{table}[t]
  \caption{Runtimes. The units is miliseconds. }
  \label{tab:exp-time}
  \centering
    \begin{tabular}{|c|c|c|c|} \hline 
       Backbone & GPU & CPU \\ \hline
      Proposed w/ ResNet18   & 1.90 & 7.70   \\
      Proposed w/ ResNet34   & 2.06 & 11.73  \\
      Proposed w/ ResNet50   & 2.52 & 22.78  \\
      Proposed w/ EffNet B0  & 2.31 & 13.59  \\
      EfficientDet D0        & 3.89 & 33.52    \\
      \hline
    \end{tabular}
\end{table}

\section{Experiments}
\label{s:exp}

To investigate how well the automatic inspection algorithm developed in this study can predict the presence or absence of defects and how accurately locate them, we conducted an evaluation experiment using the dataset described in Section~\ref{s:dataset}. 

\textbf{Backbone: }
The type of the backbone in the proposed framework can be selected. 
We examined four types of backbones: EfficientNet~\cite{tan2019efficientnet}, ResNet18~\cite{he2016deep}, ResNet34, and ResNet50.
The reason for choosing EfficientNet is to compare it with EfficientDet. 
In Section~\ref{s:model} we mentioned that it is preferable to simplify the model structure for the LUVT image analysis. 
EfficientDet is EfficientNet as its backbone and many additional modules for generic object detection. 
EfficientNet has several options for the model size. 
In this experiment, B0 was chosen. 
The corresponding option for EfficientDet is D0. 
EfficientNet and EfficientDet shall be abbreviated to EffNet and EffDet, respectively, at the lack of space.  
Combining the proposed framework with EfficientNet and comparing it with EfficientDet allows us to verify our idea discussed in Section~\ref{s:model}. 
In addition, we added typical convolutional networks, ResNet18, ResNet34, and ResNet50, to test the performance when the backbone is changed to other convolutional networks.

\textbf{Computational resources: }
Two types of computers were used. 
One was equipped with a GPU, and the other was not, as summarized in Table~\ref{tab:exp-mem} and Table~\ref{tab:exp-spec-a}. 
Both yield the same prediction results, but with different computation times. 
The GPU machine makes faster prediction, but consumes more power. 
Laptops with GPUs exist, but they are too heavy currently, increasing the burden on the human inspector.
The other computer does not have a GPU, and thus provides a more realistic environment for implementing the proposed method.

\textbf{Performance assessment: }
A positive example in the testing dataset was treated as a true positive if the classification head predicted that the defect was present and the distance between the predicted position and the correct defect position in the regression head was within $r$ pixels, where $r$ is a pre-specified error margin. 
A positive example was a false negative if no defect was predicted or the distance between predicted position and the true position was over $r$ pixels. 

\textbf{Comparison of different backbones: }
Figure~\ref{fig:acc} plots the precision and the recall for different error margins $r$.
For a threshold value of $r \ge 80$, both the differences in the two performance measures of the four proposed methods were small (At $r=80$, the difference in the precision was 0.005 and the difference in the recall was 0.023 in case of comparing EfficientNet with ResNet18). 

When the threshold $r$ was too small, neither method predicted correctly.
As the threshold was gradually increased, the precision and the recall of the proposed method with ResNet18 went up earlier than those with EfficientNet.
For example, at $r=40$, both Precision and Recall for ResNet18 were $0.92$, while they were $0.66$ and $0.67$ for EfficientNet.
Around $r=30$, the precision and the recall were lower than those of ResNet18 when the larger modules ResNet34 was used as the backbone. 
This is due to overfitting caused by the excessive capacity of the backbone.
ResNet34 is a module with a larger capacity than ResNet18. 
ResNet34 is 1.90 times larger than ResNet18. 
Curiously, however, ResNet50, which has a larger capacity, achieved higher precision and recall than ResNet34.
Balestriero et al. reported that large capacities can lead to overfitting, although further increasing the capacity can once again lower the generalization error~\cite{Randall22}, which might explain this phenomenon. 

\textbf{Comparison with EfficientDet: }
We compared the performance of the proposed model and EfficientDet where EfficientDet was a conventional state-of-the-art model.
The precision and the recall of EfficientDet are plotted against the error margin $r$ in Figure~\ref{fig:acc}. 
The detection performance of the proposed method exceeded that of EfficientDet when $r\ge 40$, which means that the proposed model reduced false detections and missed defects of the conventional method.
There are some qualitative differences between the defect detection results of EfficientDet and the proposed method.
Many defect-free images in the datast were contaminated with considerable measurement noise and other factors that caused waves to be disturbed regardless of defects.
The prediction results for such defect-free images are shown in Figure~\ref{fig:predimg}(a),(b).
EfficientDet often over-detected such wave disturbances, bringing many false positives. 
Another typical case was that the scattered waves emitted from the defect were momentarily missing after the waves reached the defect, and the defect was not clearly visible.
The prediction results for images with such defects are shown in Figure~\ref{fig:predimg}(c),(d).
In many cases, the proposed method correctly detected the defects, whereas EfficientDet often missed such defects.

It can be seen from Figure~\ref{fig:acc} that the curves of the precision and the recall of EfficientDet rise faster than those of the proposed model. 
This suggests that EfficientDet has a smaller error in estimating the location of detected defects.
The prediction results for a certain defect image are shown in Figure~\ref{fig:predimg}(e),(f).
It is observed that EfficientDet often estimated the location of defects more accurately when they were detected correctly.
However, the inspection is not entirely left to the machine in automated ultrasonic non-destructive testing using LUVT.  
Defects detected by the machine may be reconfirmed by the human inspector. 
Hence, a certain amount of error in the prediction of defect positions is tolerated.
The curves of $r\ge 80$ in Figure~\ref{fig:acc} suggests that, without demanding a too small localization error, the proposed methods can achieve much higher precision and recall than EfficientDet. 

\textbf{Comparison of computation time: }
We investigated whether the increased number of parameters compared to EfficientDet, a conventional object detection model, would result in a practical problem in terms of computation time required for prediction.
The results are shown in Table~\ref{tab:exp-time}.
EfficientDet took 3.89 msec per image using a GPU and 33.52 msec using a CPU.
The proposed method using ResNet18 required 1.90 msec of GPU time and 7.70 msec of CPU time.
The runtime was reduced despite the increase in capacity by a factor of more than three. 
This could be attributed to the elimination of unnecessary modules from the conventional object detection model.

\section{Conclusions}
In this paper, we proposed a deep framework for automatic defect detection and location estimation in LUVT.
In order to explore the structure of a neural network suitable for this task, the differences between the LUVT analysis problem and the generic object detection problem were elucidated. 
Then, the network was reconfigured by examining the necessity of each module in a state-of-the-art model for generic object detection. 
Computer experiments using actual measurement data of SUS304 flat plates showed that the prediction performance of the proposed methods were superior to that of the conventional object detection method.
The experimental results also showed that the computational time required for prediction is faster than that of the state-of-the-art model.
In the future, we plan to investigate the effectiveness of the proposed method for other types of materials such as CFRP generating anisotropic scatter waves.

\section*{Declaration of Competing Interest}
The authors declare that they have no known competing financial
interests or personal relationships that could have appeared to influence
the work reported in this paper.

\section*{Acknowledgment}
This work was supported by SECOM Science and Technology Foundation and JSPS KAKENHI (C)(21K0423100).
This work used computational resources provided by Kyoto University and
Hokkaido University through Joint
Usage/Research Center for Interdisciplinary Large-scale Information
Infrastructures and High Performance Computing Infrastructure in Japan
(Project ID: jh220033).




\bibliographystyle{elsarticle-num}
\bibliography{jiii2023miya-draft-bib,kato2205,nakajmiya-bib}



\end{document}